\documentclass[journal,twoside]{IEEEtran}

\usepackage{graphicx}
\usepackage{amsmath,amssymb} 
\usepackage{color}
\usepackage{balance}
\usepackage{times}
\usepackage{bm}
\usepackage{epsfig}
\usepackage{graphicx}
\usepackage{amsmath}
\usepackage{amssymb}
\usepackage{algorithm}
\usepackage{algpseudocode}
\usepackage[tight,footnotesize]{subfigure}
\usepackage{array}

\makeatletter
\newcommand{\thickhline}{%
    \noalign {\ifnum 0=`}\fi \hrule height 1pt
    \futurelet \reserved@a \@xhline
}

\newcommand{\thinline}{%
    \noalign {\ifnum 0=`}\fi \hrule height 0.5pt
    \futurelet \reserved@a \@xhline
}

\newcolumntype{"}{@{\vrule width 1.1pt}}
\newcolumntype{;}{@{\vrule width 1.1pt}}
\makeatother

\ifCLASSINFOpdf
\else
\fi

\begin{document}
%
\title{Adaptive Nonparametric Image Parsing}
%
%
%

\author{Tam~V.~Nguyen,
        Canyi~Lu,
        Jose~Sepulveda,
        and~Shuicheng~Yan,~\IEEEmembership{Senior~Member,~IEEE}
\thanks{Manuscript received XX XX, XXXX; revised XX XX, XXXX, and XX XX, XXXX, accepted XX XX, XXXX. This work was supported by the Singapore Ministry of Education under Grants MOE2012-TIF-2-G-016 and MOE2014-TIF-1-G-007. This paper was recommended by Associate Editor C. Shan.}
\thanks{T. Nguyen and J. Sepulveda are with the Department for Technology, Innovation and Enterprise, Singapore Polytechnic, Singapore 139651, e-mail: \{nguyen\_van\_tam, sepulveda\_jose\}@sp.edu.sg.}
\thanks{C. Lu and S. Yan are with the Department of Electrical and Computer Engineering, National University of Singapore, Singapore 119077, email: \{canyilu, eleyans\}@nus.edu.sg.}
\thanks{Color versions of one or more of the figures in this paper are available
online at http://ieeexplore.ieee.org.}
}

%
%

\markboth{JOURNAL,~Vol.~XX, No.~X, December~XXXX}%
{Nguyen \MakeLowercase{\textit{et al.}}: Adaptive Nonparametric Image Parsing}
%



\maketitle
\begin{abstract}
In this paper, we present an \textit{adaptive} nonparametric solution to the image parsing task, namely annotating each image pixel with its corresponding category label. For a given test image, first, a \textit{locality-aware} retrieval set is extracted from the training data based on super-pixel matching similarities, which are augmented with feature extraction for better differentiation of local super-pixels. Then, the category of each super-pixel is initialized by the majority vote of the $k$-nearest-neighbor super-pixels in the retrieval set. Instead of fixing $k$ as in traditional non-parametric approaches, here we propose a novel \textit{adaptive} nonparametric approach which determines the sample-specific $k$ for each test image. In particular, $k$ is adaptively set to be the number of the fewest nearest super-pixels which the images in the retrieval set can use to get the best category prediction. Finally, the initial super-pixel labels are further refined by contextual smoothing. Extensive experiments on challenging datasets demonstrate the superiority of the new solution over other state-of-the-art nonparametric solutions.
\end{abstract}


\IEEEpeerreviewmaketitle

\section{Introduction}
\IEEEPARstart{I}{mage} parsing, also called scene understanding or scene labeling, is a fundamental task in computer vision literature \cite{Gould,Kumar,Lempitsky,LiuCe,TigheECCV2010,Heesoo2,HeZemel,LabelMe,Rabinovich,Galleguillos,HeesooTensor,Munoz}.  However, image parsing is very challenging since it implicitly integrates the tasks of object detection, segmentation, and multi-label recognition into one single process. Most current solutions to this problem follow the two-step pipeline. First, the category label of each pixel is initially assigned by using a certain classification algorithm. Then, contextual smoothing is applied to enforce the contextual constraints among the neighboring pixels. The algorithms in the classification step can be roughly divided into two categories, namely parametric methods and nonparametric methods.

\begin{figure*}[!t]
\centering
\includegraphics[width = \linewidth]{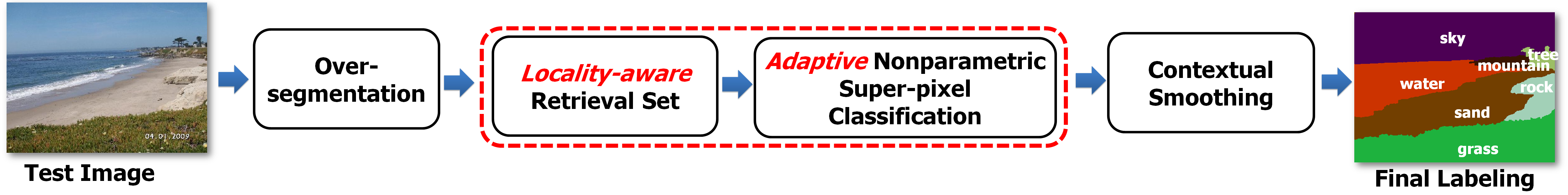}
\caption{The flowchart of our proposed nonparametric image parsing. Given a test image, we segment the image into super-pixels. Then the locality-aware retrieval set is extracted by using super-pixel matching, and the initial category label of each super-pixel is assigned by adaptive nonparametric super-pixel classification. The initial labels, in combination with contextual smoothing, give a dense labeling of the test image. The red rectangle highlights the new contributions of this work, and removing the keywords of \textit{locality-aware} and \textit{adaptive} in red then leads to the traditional nonparametric image parsing pipeline.} 
\label{fig:framework}
\end{figure*}

\textbf{Parametric methods \ }  
Fulkerson et al. \cite{Fulkerson} constructed an SVM classifier on the bag-of-words histogram of local features around each super-pixel. Tighe et al. \cite{TigheCVPR2013} combined super-pixel level features with per-exemplar sliding window detectors to improve the performance. Socher et al. \cite{Socher} proposed a method to aggregate super-pixels in a greedy fashion using a trained scoring function. The originality of this approach is that the feature vector of the combination of two adjacent super-pixels is computed from the feature vectors of the individual super-pixels through a trainable function. Farabet et al. \cite{FarabetICML} later proposed to use a multiscale convolutional network trained from raw pixels to extract dense feature vectors that encode regions of multiple sizes centered at each pixel. 


\textbf{Nonparametric methods \  }Different from parametric methods, nonparametric or data-driven methods liaise with $k$-nearest neighbors classifiers \cite{LiuCe,TigheECCV2010}. Liu et al. \cite{LiuCe} proposed a nonparametric image parsing method based on estimating SIFT Flow, a dense deformation field between images.  Given a test and a training image, the annotated category labels of the training pixels are transferred to the test ones via pixel correspondences. However, inference via pixel-wise SIFT Flow is currently very complex and computationally expensive. Therefore, Tighe et al. \cite{TigheECCV2010}  further transferred labels at the level of super-pixels, or coherent image regions produced by a bottom-up segmentation method. In this scheme, given a test image, the system searches for the top similar training images based on global features. The super-pixels of the most similar images are obtained as a retrieval set. Then the label of each super-pixel in the test image is assigned based on the corresponding $k$  most similar super-pixels in the retrieval set. Eigen et al. \cite{Eigen} further improved \cite{TigheECCV2010} by learning per-descriptor weights that minimize classification error. In order to improve the retrieval set, Singh et al. \cite{Singh} used adaptive feature relevance and semantic context. They adopted a locally adaptive distance metric which is learned at query time to compute the relevance of individual feature channels. Using the initial labelling as a contextual cue for presence or absence of objects in the scene, they proposed a semantic context descriptor which helped refine the quality of the retrieval set. In a different work, Yang et al. \cite{Yang} looked into the long-tailed nature of the label distribution. They expanded the retrieval set by rare class exemplars and thus achieved more balanced super-pixel classification results. Meanwhile, Zhang et al.~\cite{Honghui} proposed a method which exploits partial similarity between images. Namely, instead of retrieving global similar images from the training database, they retrieved some partially similar images so that for each region in the test image, a similar region exists in one of the
retrieved training images.

Due to the limited discriminating power of classification algorithms, the output initial labels of pixels may be noisy. To further enhance the label accuracy, contextual smoothing is generally used to exploit global contexts among the pixels. Rabinovich et al. \cite{Rabinovich} incorporated co-occurrence statistics of category labels of super-pixels into the fully connected Conditional Random Field (CRF). Galleguillos et al. \cite{Galleguillos} proposed to exploit the information of relative location such as above, beside, or enclosed between super-pixel categories. Meanwhile, Myeong et al. \cite{Heesoo2} introduced a context link view of contextual knowledge, where the relationship between a pair of annotated super-pixels is represented as a context link on a similarity graph of regions, and link analysis techniques are used to estimate the pairwise context scores of all pairs of unlabeled regions in the input image. Later, \cite{HeesooTensor} proposed a method to transfer high-order semantic relations of objects from annotated images to unlabeled images. Zhu et al. \cite{Long} proposed the hierarchical image model composed of rectangular regions with parent-child dependencies. This model captures large-distance dependencies and is solved efficiently using dynamic programming. However, it supports neither multiple hierarchies, nor dependencies between variables at the same level. In another work, Tu et al. \cite{Tu} introduced a unified framework to pool the information from segmentation, detection and recognition for image parsing. They have to spend much effort to design such complex models. Due to the complexity, the proposed model might not scale well with different datasets. 

In this work, our focus is placed on nonparametric solutions to the image parsing problem. However, there are several shortcomings in existing nonparametric methods. First, it is often quite difficult to get globally similar images to form the retrieval set. Also by only considering global features, some important local components or objects may be ignored. Second, $k$ is fixed empirically in advance in such a nonparametric image parsing scheme. Tighe et al. \cite{TigheECCV2010} reported the best results by varying $k$ on the test set. However, this strategy is impractical since the ground-truth labels are not provided in the testing phase. Therefore, the main issues in the context of the nonparametric image parsing are 1) how to get a good retrieval set, and 2) how to choose a good $k$ for initial label transfer. In this work, we aim to improve both aspects, and the main contributions of this work are two-fold. \begin{enumerate}
\item Unlike the traditional retrieval set which consists of globally similar images, we propose the \textit{locality-aware} retrieval set. The \textit{locality-aware} retrieval set is extracted from the training data based on super-pixel matching similarities, which are augmented with feature extraction for better differentiation of local super-pixels. 

\item Instead of fixing $k$ as in traditional nonparametric methods, we propose an \textit{adaptive} method to set the sample-specific $k$ as the number of the fewest nearest neighbors which similar training super-pixels can use to get their best category label predictions.
 
\end{enumerate}

\section{Adaptive Nonparametric Image Parsing}
\label{sec:framework}

\subsection{Overview}
Generally, for nonparametric solutions to the image parsing task, the goal is to label the test image at the pixel level based on the content of the retrieval set, but assigning labels on a per-pixel basis as in \cite{LiuCe,FarabetICML} would be too inefficient. In this work, we choose to assign labels to super-pixels produced by bottom-up segmentation as in~\cite{TigheECCV2010}. This not only reduces the complexity of the problem, but also gives better spatial support for aggregating features belonging to a single object than, say, fixed-size square patches centered at each pixel in an image. 

The training images are first over-segmented into super-pixels by using the fast graph-based segmentation algorithm of \cite{Superpixels} and their appearances are described using 20 different features similar to those of \cite{TigheECCV2010}. The complete list of super-pixel's features is summarized in Table \ref{table:features}. Each training super-pixel is assigned a category label if 50\% or more of the super-pixel overlaps with a ground truth segment mask of that label. For each super-pixel, we perform feature extraction and then reduce the dimension of the extracted feature. 

\begin{table}[!t]
\caption{The list of all super-pixel's features.}
\begin{center}
\begin{tabular}{|l|c |l|c|}
\hline
\textbf{Type} & \textbf{Dim}& \textbf{Type} & \textbf{Dim}\\

\hline
Centered mask & 64 &	SIFT histogram top&100\\
\hline
Bounding box	& 2 &	SIFT histogram right &100\\
\hline
Super-pixel area	& 1 &	SIFT histogram left &100\\
\hline
Absolute mask	& 64 &	Mean color&3\\
\hline
Top height 	& 1 &	Color standard deviation&3\\
\hline
Texton histogram	& 100 &	Color histogram&33\\
\hline
Dilated	texton histogram & 100 &	Dilated color histogram&33\\
\hline
SIFT histogram	& 100 &	Color thumbnail&192\\
\hline
Dilated SIFT histogram	& 100 &	Masked color thumbnail &192\\
\hline
SIFT histogram bottom	& 100 &	GIST&320\\
\hline
\end{tabular}
\end{center}

\label{table:features}
\end{table}

For the test image, as illustrated in Figure \ref{fig:framework}, over-segmentation and super-pixel feature extraction are also conducted. Next, we perform the super-pixel matching process to obtain the locality-aware retrieval set. The adaptive nonparametric super-pixel classification is proposed to determine the initial label of each super-pixel. Finally, the graphical model inference is performed to preserve the semantic consistency between adjacent pixels. More details of the proposed framework, namely the locality-aware retrieval set, adaptive nonparametric super-pixel classification, and contextual smoothing, are elaborated as follows.

\begin{figure*}[!t]
\centering
\includegraphics[width = \linewidth]{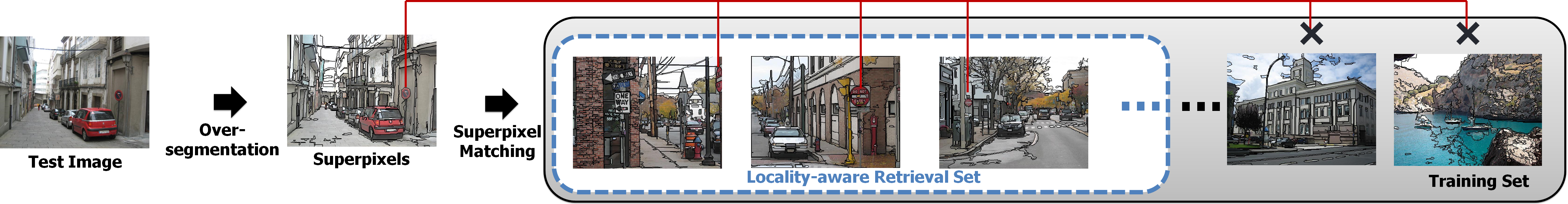}
\caption{The process to extract the retrieval set by super-pixel matching. The test image is first oversegmented into super-pixels. Then, we compute the similarity between the test image and each training image as described in Algorithm~\ref{alg:retrieval}. (Please view in high 400\% resolution).}
\label{fig:matching}
\end{figure*}

\subsection{Locality-aware Retrieval Set}

For nonparametric image parsing, one important step of parsing a test image is to find a retrieval set of training images that will serve as the reference of candidate super-pixel level annotations. This is done not only for computational efficiency, but also to provide scene-level context for the subsequent processing steps. A good retrieval set should contain images of a similar scene type as that of the test image, along with similar objects and spatial layouts. Unlike \cite{TigheECCV2010} where global features are used to obtain the retrieval set, we utilize the super-pixel matching as illustrated in Figure \ref{fig:matching}. The motivation is that sometimes it may be difficult to get globally similar images, especially when the training set is not big enough, yet locally similar ones are easier to obtain; also sometimes if only global features are considered for retrieval set selection, some important local components or objects may be ignored. In this work, the retrieval set is selected based on local similarity measured over super-pixels. To enhance the discriminating power of super-pixels, we utilize Linear Discriminant Analysis (LDA) \cite{LDA} for feature reduction to a lower feature  dimension. Then we use the augmented super-pixel similarity instead of global similarity to extract the retrieval set.

Denote $x \in \mathbb{R}^{n_x \times 1}$ as the original feature vector of the super-pixel, where $n_x$ is the dimension of the feature vector. The corresponding feature vector $\hat{x}$ after the feature reduction is computed as,
\begin{equation}
\hat{x} = \bm{W}x,
\end{equation}
where $\bm{W}$ is the transformation matrix. In particular, LDA looks for the directions that are most effective for discrimination by minimizing the ratio between the intra-category ($\bm{S_w}$) and inter-category ($\bm{S_b}$) scatters:

\begin{equation}
\bm{W}^{*} = \arg \min_{\bm{W}}\frac{|\bm{W}^{T}\bm{S_w}\bm{W}|}{|\bm{W}^{T}\bm{S_b}\bm{W}|},
\end{equation}

\begin{equation}
\bm{S_w} = \sum_{i=1}^N(\bm{x_i - \bar{x}^{c_i}})(\bm{x_i - \bar{x}^{c_i}})^{T},
\end{equation}

\begin{equation}
\bm{S_b} = \sum_{c=1}^{N_c} n_c(\bm{\bar{x}^c - \bar{x}})(\bm{\bar{x}^c - \bar{x}})^{T},
\end{equation}\\
where $N$ is the number of super-pixels in all training images, $N_c$ is the number of categories, $n_c$ is the number of super-pixels for the $c$-th category, $\bm{x}_i$, $\forall i \in \{1,\cdots,N\}$, is the feature vector of one training super-pixel, $c_i$ is the category label of the $i$-th super-pixel in the training images, $\bm{\bar{x}}$ is the mean of feature vector of training super-pixels, and $\bm{\bar{x}^c}$ is the mean of the  $c$-th category. Note that the category label of each super-pixel is obtained from the ground-truth object segment with the largest overlapping with the super-pixel. As shown in \cite{LDA}, the projection matrix $\bm{W}^{*}$ is composed of the eigenvectors of $\bm{S_w^{-1}}\bm{S_b}$. Note that there are at most $N_c - 1$ eigenvectors with non-zero real corresponding eigenvalues since there are only $N_c$ points to compute $\bm{S_b}$. In other words, the dimensionality of $\bm{W}$ is $N_c - 1 \times n_x$. Therefore, LDA naturally reduces the feature dimension to $N_c$$-$$1$ in the image parsing task. Since the category number is much smaller than the feature number, the benefits of the reduced dimension include the shrinkage of memory storage and the removal of those less informative features for consequent super-pixel matching. Obviously the reduction of feature dimension is also beneficial to the nearest super-pixel search in the super-pixel classification stage.

\begin{algorithm}[!t]
\small
  \caption{Locality-aware Retrieval Set Algorithm}\label{alg:retrieval}
  \begin{algorithmic}[1]
	\State \textbf{parameters:}
	$n_q$, $\bm{n_t}$, $N_I$, $\bm{Q}$, $\bm{T}$.
	\State The unique index set $S = \emptyset$.
	
	\State $\bm{v}=\bm{0}\in\mathbb{R}^{N_I}$.
    \For{\texttt{i = 1:$n_q$}}
    	\State [$\bm{\eta}_i$, $\bm{\Delta}_i$]$\leftarrow$\textsc{Knn}($\bm{Q}_i$, $\bm{T}$, $k_m$);
    	\State $\bm{\eta}_i\leftarrow\bm{\eta}_i \verb|\| S$;
		\If{$\bm{\eta}_i \neq \emptyset$}
    	\State $\bm{\eta}_i$$\leftarrow$\textsc{\textbf{RefineIndexSet}}($\bm{\eta}_i$);
    	\State $\bm{I}_i\leftarrow$\textsc{\textbf{FindImageIndex}}($\bm{\eta}_i$);
    	\State $\bm{v}(\bm{I}_i)\leftarrow \bm{v}(\bm{I}_i)+1./\bm{\Delta}_i(\bm{\eta}_i)$;
       	\State $S\leftarrow S\bigcup \bm{\eta}_i$;
       	\EndIf
	\EndFor
	\State $\bm{v}=$\textsc{\textbf{NormalizeAndSort}}($\bm{v}$).
	\State $k_r = \arg \min_{u}\frac{\sum_{j=1}^{u}v_j}{\sum_{j=1}^{N_I}v_j} \geq \tau$.
	\State \textbf{return} top $k_r$ training images.
	\item[]
	\Function{RefineIndexSet}{$\eta$, $\Delta$}
	\State $\bm{d}=\bm{\infty}\in\mathbb{R}^{N_I}$.
	\State $\bm{\Gamma} = \emptyset$.
	\For{\texttt{i = 1:$|\eta|$}}
		\If{$\bm{d}(\bm{m}(\eta_i)) > \Delta_i$ }
			\State $\bm{d}(\bm{m}(\eta_i)) = \Delta_i$; 
		\Else
			\State $\bm{\Gamma} = \bm{\Gamma} \bigcup \texttt{i}$;
		\EndIf
	\EndFor	
	\State \textbf{return} $\bm{\Gamma}$.
	\EndFunction
	\item[]
	\Function{FindImageIndex}{$\eta$}
		\State $\bm{\Gamma}=\bm{\infty}\in\mathbb{R}^{|\eta|}$.
		\For{\texttt{i = 1:$|\eta|$}}
			\State $\Gamma_i$ = $\bm{m}(\eta_i)$;
		\EndFor	
		\State \textbf{return} $\Gamma$.	
	\EndFunction
	\item[]
	\Function{NormalizeAndSort}{$\bm{v}$}
		\State $\bm{\Gamma}=\bm{\infty}\in\mathbb{R}^{|\bm{v}|}$.
		\For{\texttt{i = 1:$|\bm{v}|$}}
			\State $\Gamma_i$ = $\bm{v}_i/\min{(n^t_i, n_q)}$;
		\EndFor	
		\State $\Gamma = \mathrm{sort}(\Gamma)$. 
		\State \textbf{return} $\Gamma$.	
	\EndFunction
  \end{algorithmic}
\end{algorithm}

\begin{figure*}[!t]
\centering
\includegraphics[width = 0.8\linewidth]{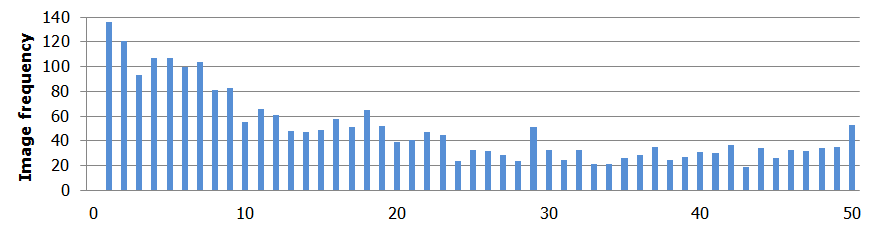}
\caption{The distribution of best $k$s for all training images in the SIFTFlow dataset. It can be observed that there is no dominant $k$ from $1$ to $50$.}
\label{fig:kqp}
\end{figure*}

The procedure to obtain the retrieval set is summarized in Algorithm \ref{alg:retrieval}. Denote $n_q$ as the number of super-pixels in the test image, ${n^t_j}\in\mathbb{R}$ as the number of super-pixels for the $j$-th training image, and $N_I$ as the number of training images. We impose the nature constraint that one super-pixel in a training image is matched with only one super-pixel of the test image. We denote $S$ as the unique index set which stores the indices of the already matched super-pixels, $\bm{v}$ as the similarity vector between the test image and all training images, $\bm{Q} \in \mathbb{R}^{(N_c-1)\times n_q}$ as the feature matrix for all the super-pixels in the test image, $\bm{T} \in \mathbb{R}^{(N_c-1)\times (\sum_{j}n^t_j)}$ as the feature matrix for all the super-pixels in the training set, and $\bm{m} \in \mathbb{R}^{\sum_{j}n^t_j}$ as the mapping index between the super-pixel and the corresponding training image. As aforementioned, the over-segmentation over the image is performed by using \cite{Superpixels}. Then we extract the corresponding features similarly as \cite{TigheECCV2010} for each super-pixel and use LDA to reduce the feature dimension. 

We match each super-pixel in the test image with all super-pixels in the training set. In order to reduce the complexity, we perform \textsc{Knn} to find the nearest $k_m$ super-pixels in the training images for the $i$-th super-pixel in the test image. The Euclidean distance is used to calculate the dissimilarity between two super-pixels. As a result, we have $\bm{\eta}_i \in \mathbb{R}^{k_m}$ as the indices of the returned nearest super-pixels of the $i$-th test super-pixel, and $\bm{\Delta}_i\in \mathbb{R}^{k_m}$ as the corresponding distances of the returned nearest super-pixels to the $i$-th test super-pixel. We remove the super-pixels in $S$ from $\bm{\eta}_i$, where $S$ includes the training super-pixels matched by the first $i$$-$$1$ test super-pixels. There may be more than one super-pixel from one training image, thus  \textsc{\textbf{RefineIndexSet}} is performed to keep the nearest one. Note that $|\cdot|$ denotes the number of the elements in an array. Then, the index set $S$ is updated by adding $\eta_i$. 

The function \textsc{\textbf{FindImageIndex}} is invoked to retrieve the corresponding image index of $\bm{\eta}_i$. Then we update the similarity vector $\bm{v}$  since the number of super-pixels is not the same for every image. For example, the number of super-pixels of SIFTFlow training set varies from $5$ to $193$. Therefore we perform \textsc{\textbf{NormalizeAndSort}} to obtain the final similarity vector. Namely, for each training image $j$, $v_j$ is divided by $\min(n_q,{n^t_j})$.  The retrieval set then includes the top $k_r$ training images by $\frac{\sum_{j=1}^{k_r}v_j}{\sum_{j=1}^{N_I}v_j} \geq \tau $, where the parameters $k_m$ and $\tau$ are selected by the grid search over the training set based on the leave-one-out strategy. Namely, we choose a pair of $\tau \in \{0.1, ..., 0.5\}$ with step size $0.1$, and $k_m \in \{500, ..., 2500\}$ with step size $500$ and perform the following adaptive non-parametric super-pixel classification for all images in the training set. The leave-one-out strategy means that when one training image is selected as a test image, the rest of training images is used as the corresponding training set.

\subsection{Adaptive Nonparametric Super-pixel Classification}

Adaptive nonparametric super-pixel classification aims to overcome the limitation of the traditional $k$-nearest neighbor ($k$-NN) algorithm which usually assigns the same number of nearest neighbors for each test sample. For nonparametric algorithms, the label of each super-pixel in the test image is assigned based on the corresponding similar $k$ super-pixels in the retrieval set. Our improved $k$-NN algorithm focuses on looking for the suitable $k$ for each test sample. 

Basically the sample-specific $k$ of each test image is propagated from its similar training images. In particular, each training image $t$ retrieved by the super-pixel matching process, is considered as one test image, while the left $N_I$$-$$1$ images in the training set are referred to the corresponding training set. Then we perform super-pixel matching to obtain the retrieval set for $t$ and assign the label $l^k_i$ of the $i$-th super-pixel by the majority vote of the $k$ nearest super-pixels in the retrieval set,
\begin{equation}
l_i^* = \arg \max_{l_i} L(k,l_i), 
\end{equation}
where $L$ is the likelihood ratio for the $i$-th super-pixel to have the category $l_i$ based on the $k$ nearest super-pixels and defined as below,
\begin{align}
L(k,l_i) = &\frac{P(i|l_i, k)}{P(i|\bar{l}_i, k)} = \frac{n(l_i,NN(i, k))/n(l_i,D)}{n(\bar{l}_i,NN(i, k))/n(\bar{l_i},D)}.
\end{align}
Here $n(l_i,NN(i, k))$ is the number of super-pixels with class label $l_i$ in the $k$ nearest super-pixels of the $i$-th super-pixel in the retrieval set, $\bar{l}_i$ is the set of all labels excluding $l_i$, and $D$ is the set of all super-pixels in the whole training set. $NN(i, k)$ consists of $k$ nearest super-pixels of the $i$-th super-pixel from the retrieval set. Then we compute the per-pixel accuracy of each retrieved training image $t$ for different $k$s. We denote $A_{tk}$ as the per-pixel performance (the percentage of all ground-truth pixels that are correctly labeled) of the training image $t$ with the parameter value $k$. We vary $k$ from $1$ to $50$ with step size $1$, $k \in \{1,~2,~3,~\cdots,~50\} $. As can be observed in Figure \ref{fig:kqp}, there is no dominant $k$ from $1$ to $50$ in the overall SIFTFlow training set. It motivates the necessity of adaptive $k$ nearest neighbors for the nonparametric super-pixel classification process. Thus, for each test image, we assign its $k$ by transferring $k$s of the similar images returned by the super-pixel matching process, 
\begin{equation}
k^* = \arg \max_k \sum_{t=1}^{k_r} A_{tk},
\end{equation}
where $k_r$ is the number of images in the retrieval set for the test image. Then based on selected $k^*$, the initial label of a super-pixel in the test image is obtained in the same way as in Eqn. (4).

\begin{figure*}[!t]
\centering
\includegraphics[width = \linewidth]{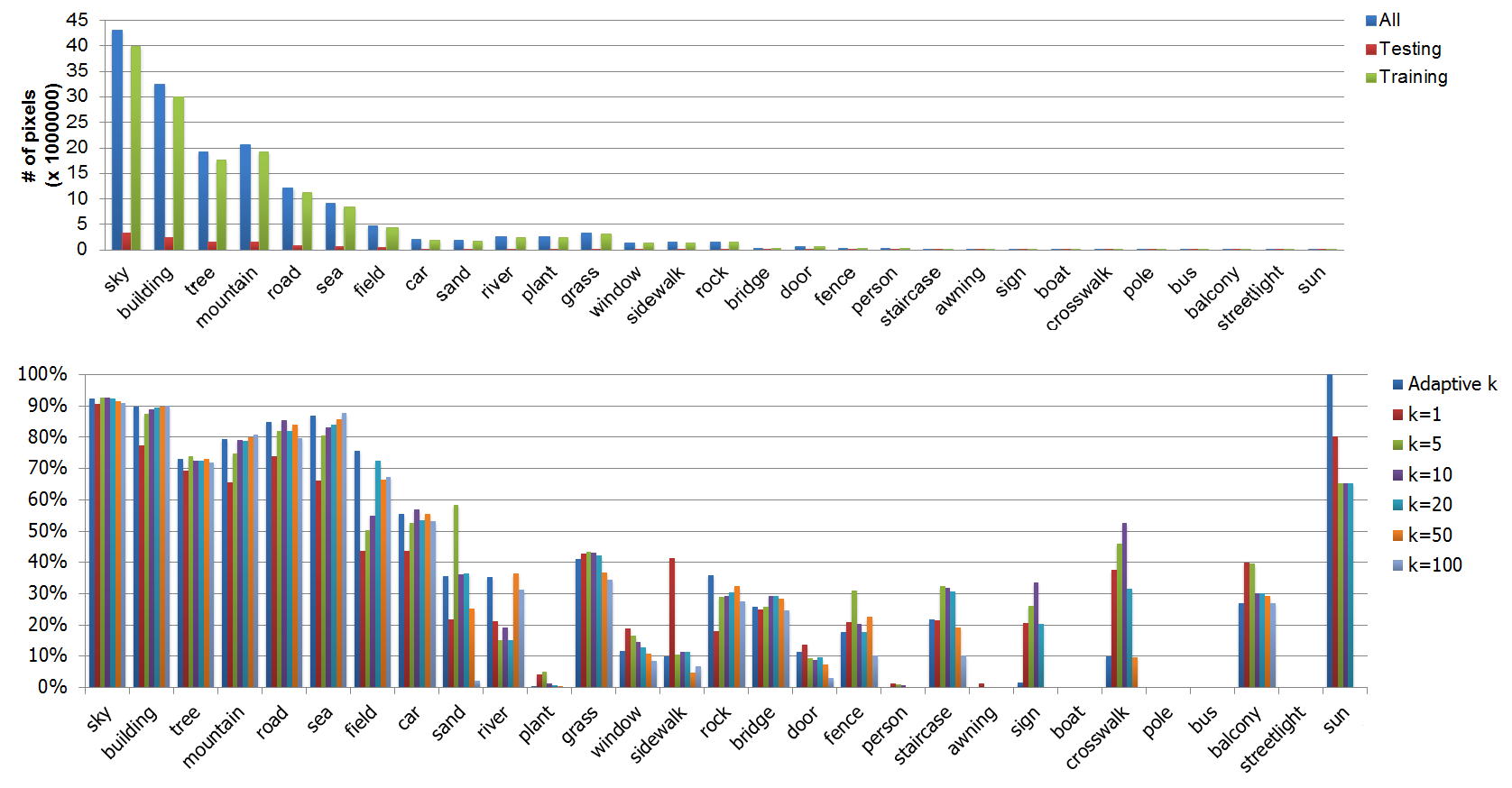}
\caption{(Top) Label frequencies for the pixels in the SIFTFlow training set. (Bottom) The per-category classification rates of different $k$s and our adaptive nonparametric method on the SIFTFlow dataset. The categories `bird', `cow', `dessert', and `moon' are dropped from the figure since they are not present in the test split. (Please view in high 400\% resolution).}
\label{fig:SIFTFlow}
\end{figure*}

\subsection{Contextual Smoothing}

Generally, the initial labels for the super-pixels may still be  noisy, and these labels need be further refined with global context information.
The contextual constraints are very important for parsing images. For example, a pixel assigned with ``car'' is likely connected with ``road''. Therefore, the initial labels are smoothened with an MRF energy function defined over the field of pixels: 

\begin{equation}
E(l) = \sum_{i \in V}E_{d}(i,l_i) + \lambda \sum_{e_{ij} \in E} E_{s}(l_i,l_j),
\end{equation}
where $V$ is the set of all pixels in the image, $E$ is the set of edges connecting adjacent pixels, and $\lambda$ is a smoothing constant. The data term is defined as follows
\begin{equation}
E_{d}(i,l_i) = -\log L(k^*, l_{sp(i)}),
\end{equation}
where $sp(i)$ means the super-pixel containing the $i$-th super-pixel and the $L$ function is defined in Eqn. (5). The MRF model also includes the smoothness constraint reflecting the spatial consistency (pixels or super-pixels close to each other are most likely to have similar labels). Therefore, the smoothing term $E_s(l_i, l_j)$ imposes a penalty when two adjacent pixels ($p_i$, $p_j$) are similar but are assigned with different labels ($l_i$, $l_j$). 
$E_{s}$ is defined based on probabilities of label co-occurrence and biases the neighboring pixels to have the same label in the case that no other information is available, and the probability depends on the
edge of the image: 
\begin{equation}
E_{s}(l_i; l_j) =  - \xi_{ij}  \times \log\left(\frac{P(l_i|l_j) + P(l_j|l_i)}{2}\right) \times \delta[l_i \neq l_j],
\end{equation}
where $P(l_i|l_j)$ is the conditional probability of one pixel having label $l_i$ given that its neighbor has label $l_j$, estimated by counts from the training set. $\xi_{ij}$ is defined based on the normalized gradient value of the neighboring pixels:
\begin{equation}
\xi_{ij} = \frac{\nabla_{ij}}{\sum_{e_{pq} \in E}\nabla_{pq}},
\end{equation}
where $\nabla_{ij} = ||I(i) - I(j)||^2$ is the $\ell_2$ norm of the gradient of the test image $I$ at a pixel $i$ and its neighbor pixel $j$. The stronger the luminance edge is, the more likely the neighboring pixels may have different labels. Multiplication with the constant Potts penalty $\delta[l_i \neq l_j]$ is necessary to ensure that this energy term is semi-metric as required by graph cut inference~\cite{MRF1}. We perform the inference using the $\alpha-\beta$ swap algorithm~\cite{MRF1,MRF2,MRF3}.

\section{Experiments}
\label{sec:experiments}
\subsection{Datasets and Evaluation Metrics}

In this section, our approach is validated on two challenging datasets: SIFTFlow \cite{LiuCe} and 19-Category LabelMe~\cite{Jain}.

\textbf{SIFTFlow dataset}\footnote{http://people.csail.mit.edu/celiu/LabelTransfer/LabelTransfer.rar} is composed of 2,688 images that have been throughly labeled by LabelMe users. The image size is 256 $\times$ 256 pixels. Liu et al. \cite{LiuCe} split this dataset into 2,488 training images and 200 test images and used synonym correction to obtain 33 semantic labels (sky, building, tree, mountain, road, sea, field, car, sand, river, plant, grass, window, sidewalk, rock, bridge, door, fence, person, staircase, awning, sign, boat, crosswalk, pole, bus, balcony, streetlight, sun, bird, cow, dessert, and moon).

\textbf{19-Category LabelMe dataset}\footnote{http://www.umiacs.umd.edu/\symbol{126}ajain/dataset/LabelMesubsetdataset.zip} Jain et al.~\cite{Jain} randomly collected 350 images from LabelMe \cite{LabelMe} with 19 categories (grass, tree, field, building, rock, water, road, sky, person, car, sign, mountain,
ground, sand, bison, snow, boat, airplane, and sidewalk). This dataset is split into 250 training images and 100 test images. 

We evaluate our approach on both sets, but perform additional analysis on the SIFTFlow dataset since it has a larger number of categories and images. In evaluating image parsing algorithms, there are two metrics that are commonly used: per-pixel and per-category classification rate. The former rates the total proportion of correctly labeled pixels, while the latter indicates the average proportion of correctly labeled pixels in each object category. If the category distribution is uniform, then the two would be the same, but this is not the case for real-world scenes. Note that for all experiments, the $\lambda$ is empirically set as $16$ in the contextual smoothing process. $k_m$ and $\tau$   are set as $1000$ and $0.3$, respectively. In all of our experiments, we use Euclidean distance metric to find the nearest neighbors.

\begin{table}[!t]
\footnotesize
\caption{ Performance comparison of our algorithm with other algorithms on the SIFTFlow dataset~\cite{LiuCe}. Per-pixel rates and average per-category rates are presented. The best performance values are marked in \textbf{bold}. }
\begin{center}
\begin{tabular}{"p{3.7cm} |c |c;}
\thickhline
\textbf{~Algorithm} & \textbf{Per-Pixel (\%)} & \textbf{Per-Category (\%)}\\
\thickhline
\hline
\thinline

\multicolumn{3}{"c;}{Parametric Baselines} \\ 
\hline

Tighe et al. \cite{TigheCVPR2013}	& 78.6	&	\textbf{39.2}	\\
\hline
Farabet et al. \cite{FarabetICML}	& 78.5 &	29.6\\
\thickhline
\hline
\thinline
\multicolumn{3}{"c;}{Nonparametric Baselines} \\ 

\hline

Liu et al. \cite{LiuCe}	& 74.8	&	--	\\
\hline
Tighe et al. \cite{TigheECCV2010}  & 76.3	 &	28.8	\\
\hline
Tighe et al. \cite{TigheECCV2010} (adding geometric information)  &76.9	 &	29.4	\\
\hline
Myeong et al. \cite{HeesooTensor}	& 76.2	&	29.6\\
\hline
Eigen et al. \cite{Eigen} &	77.1	&	32.5	\\
\hline

\multicolumn{3}{"c;}{Our Proposed Adaptive Nonparametric Algorithm} \\ 
\hline
Super-pixel Classification	& 77.2	&	34.9	\\
\hline
Contextual Smoothing	& \textbf{78.9}	&	34.0\\
\thickhline
\end{tabular}
\end{center}

\label{table:SIFTFlow}
\end{table}

\begin{table}[!t]
\footnotesize
\caption{ Performance comparison of different $k$s and our algorithm on the SIFTFlow dataset~\cite{LiuCe}. Per-pixel rates and average
per-category rates are presented.}
\begin{center}
\begin{tabular}{|"p{3.8cm} |c |c;|}
\thickhline
\textbf{Parameter} & \textbf{Per-Pixel (\%)} & \textbf{Per-Category (\%)}\\

\thickhline
\hline
\thinline
$k$ = 1& 70.2 &	31.9\\
\hline
$k$ = 5	& 76.6 &	\textbf{34.8}\\
\hline
$k$ = 10	& 77.5 &	34.6\\
\hline
$k$ = 20	& 77.8 &	33.5\\
\hline
$k$ = 30	& 77.9 &	33.3\\
\hline
$k$ = 40	& 77.9 &	30.6\\
\hline
$k$ = 50	& 77.8 &	29.5\\
\hline
$k$ = 60	& 77.5 &	28.6\\
\hline
$k$ = 70	& 77.8 &	28.5\\
\hline
$k$  = 80	& 77.5 &	28.2\\
\hline
$k$  = 90	& 77.1 &	27.2\\
\hline
$k$  = 100	& 76.9	&	26.8	\\
\thickhline
\hline
\thinline
\textbf{Adaptive $k$ in Our Algorithm}	& \textbf{78.9}	&	34.0\\
\thickhline
\end{tabular}
\end{center}
\label{table:K}
\end{table}

\subsection{Performance on the SIFTFlow Dataset}

\textbf{Comparison of our algorithm with state-of-the-arts \ } Table \ref{table:SIFTFlow} reports per-pixel and average per-category rates for image parsing on the SIFTFlow dataset. Even though the nonparametric methods are our main baselines, we still list parametric methods for reference. 
Our proposed method outperforms the baselines by a remarkable margin. We did not compare our work with \cite{Singh} and \cite{Yang} since \cite{Singh} uses a different set of super-pixel's features whereas \cite{Yang} utilizes the extra data to balance the distribution of the categories in the retrieval set. Compared with our initial super-pixel classification result, the final contextual smoothing improves overall per-pixel rates on the SIFTFlow dataset by about 1.7\%. Average per-category rates drop slightly due to the contextual smoothing on some of the smaller classes. Note that Tighe et al. \cite{TigheCVPR2013} improved \cite{TigheECCV2010} by adding extensively multiple detectors (their performance reaches $78.6\%$). The addition of many object detectors brings a better per-category performance but also increases the processing time since running object detection is very time-consuming. Note that, to train the object detectors, \cite{TigheCVPR2013} must use extra data. Also,  \cite{TigheCVPR2013} utilizes SVM instead of $k$-NN as in our work that may bring better classification results, especially for some rare categories. Meanwhile, our proposed method improves \cite{TigheECCV2010} with a simpler solution and even achieves a better performance in terms of per-pixel rate (\textbf{$78.9\%$}). Also, our method performs better than \cite{FarabetICML} which deployed heavily deep learning features.

\begin{figure*}[!t]
\centering
\includegraphics[width = \linewidth]{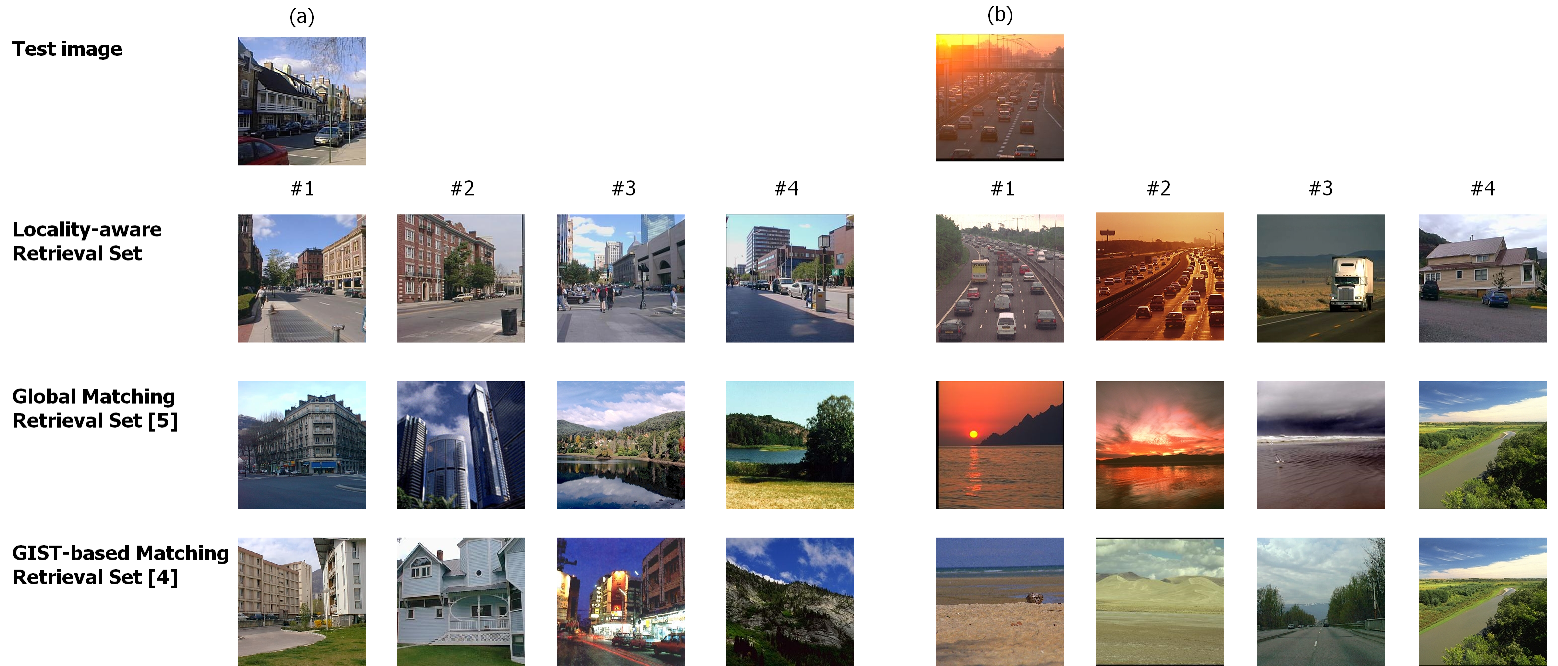}
\caption{Top 4 exemplar retrieval results of super-pixel matching, global matching~\cite{TigheECCV2010}, and GIST-based matching~\cite{LiuCe}. (a) Global matching returns ``tall building'' and ``open country'' scenes, and GIST-based matching obtains ``inside city'' and ``mountain''. Meanwhile, our method obtains the reasonable images of ``urban street''. (b) The ``open country'' images are retrieved in GIST-based matching and the ``sunset coastal'' scenes are returned in global matching instead of ``highway'' as in our method. } 
\label{fig:retrieval_res}
\end{figure*}

\begin{figure*}[!t]
\centering

\includegraphics[width = \linewidth]{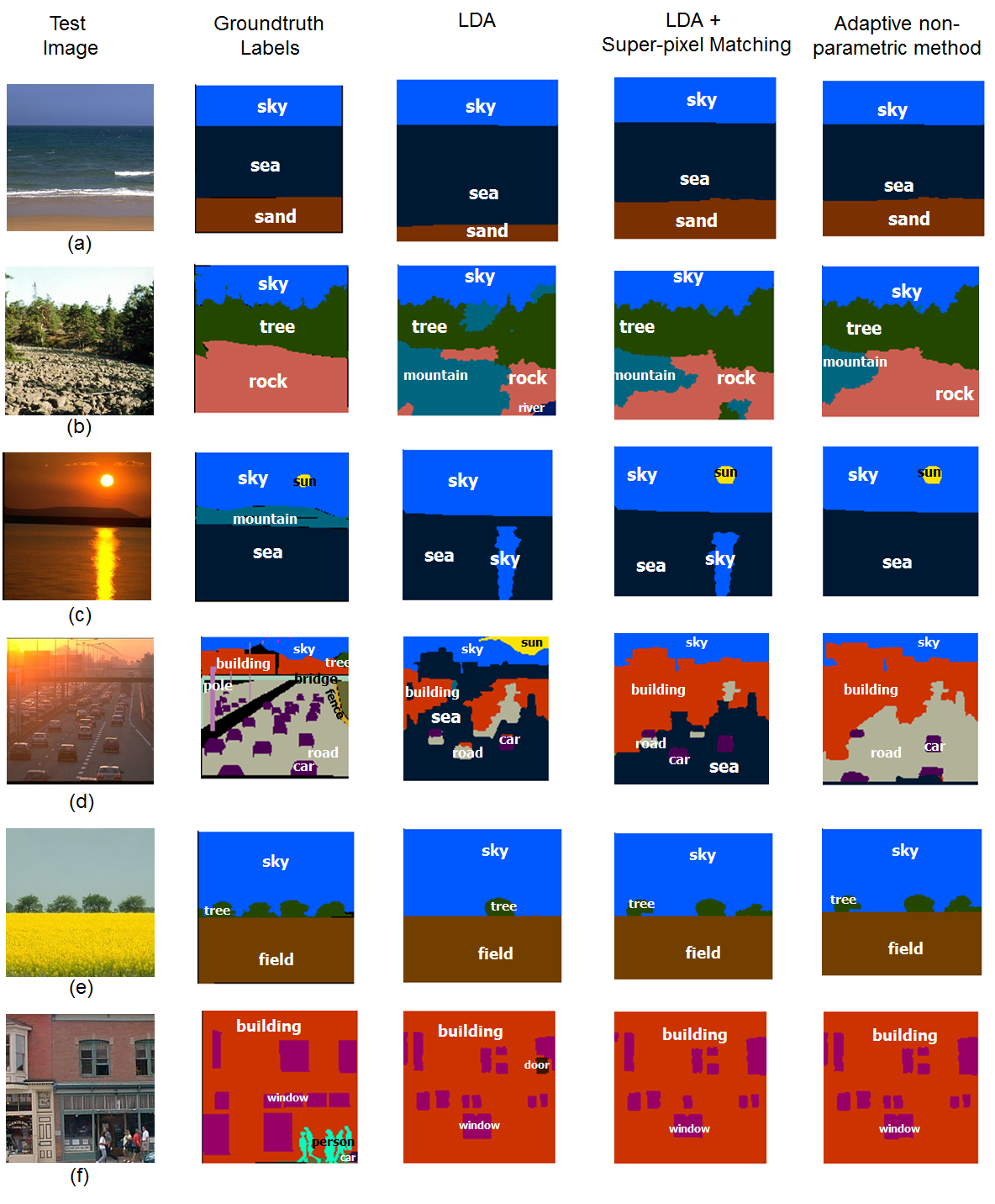}
\caption{Exemplar results from the SIFTFlow dataset. In (a), the adaptive nonparametric method successfully parses the test image. In (b), the ``rock'' is classified instead of ``river'' or ``mountain'' in other two methods. In (c), our method recovers the ``sun'' and removes the spurious classification of the sun's reflection in the water as ``sky''. In (d), the labeled ``sea'' regions in two other methods are recovered as ``road''. In (e), some of the trees are recovered in the adaptive nonparametric method. In (f), our method recovers ``window'' from ``door''. Best viewed in color. } 

\label{fig:SIFTFlow_res}
\end{figure*}

\begin{table}[!t]
\normalsize
\caption{ Performance comparison of different settings on the SIFTFlow dataset~\cite{LiuCe}. Per-pixel classification rates (with per-category rates in parentheses) are presented.}
\begin{center}
\begin{tabular}{|"p{6.4cm} |c;|}
\thickhline
\textbf{Algorithm} & \textbf{Performance}\\
\thickhline
\hline
\thinline
\multicolumn{2}{|"c;|}{Baseline} \\ 

\hline

SuperParsing \cite{TigheECCV2010}   & 76.3 (28.8)		\\

\thickhline
\hline
\thinline

\multicolumn{2}{|"c;|}{Our Improvements} \\ 
\hline

SuperParsing + LDA + Global Matching +  (fixed $k$ = 20)	&	76.4 (31.2)\\
\hline
SuperParsing + LDA + Super-pixel Matching + ($k$~=~20) 	&	77.8 (33.5)	\\
\hline
SuperParsing + LDA + Super-pixel Matching + Adaptive $k$		& \textbf{78.9 (34.0)}	\\
\thickhline

\end{tabular}
\end{center}

\label{table:Improvement}
\end{table}

\textbf{Performance of different $k$s \ }  The impact of different $k$s is further investigated on the SIFTFlow dataset. In this experiment, the parameter $k$ varies from $1$ to $100$. LDA and super-pixel matching are utilized in order to keep fair comparison with our adaptive nonparametric method. Table~\ref{table:K} summarizes the performance of different $k$s on both per-pixel and per-category criteria. The relationship between per-pixel and per-category of different $k$s is inconsistent. The smaller $k$s ($\leq 20$) tend to achieve a higher per-category whereas the larger $k$s lean to a higher per-pixel rate. A lower $k$ responds well with rare categories (i.e., boat, pole, bus, etc. as illustrated in Figure~\ref{fig:SIFTFlow}), thus it leads to improved per-category classification. Meanwhile, a higher $k$ leads to better per-pixel accuracy since it works well for more common categories such as sky, building, and tree. $k = 5$ yields the largest per-category rate, but its per-pixel performance is much lower than that of $k = 40$. As a closer look, Figure~\ref{fig:SIFTFlow} also shows the details of per-category classification rates of different $k$s. The smaller $k$s yield better results on the categories with a small number of samples while the larger $k$s are sensitive on categories with a large number of samples such as \textit{sky}, \textit{sea}, etc. As observed in the same Figure \ref{fig:SIFTFlow}, our adaptive nonparametric approach exhibits advantages over smaller and larger $k$s.

\begin{figure*}[!t]
\centering
\includegraphics[width = \linewidth]{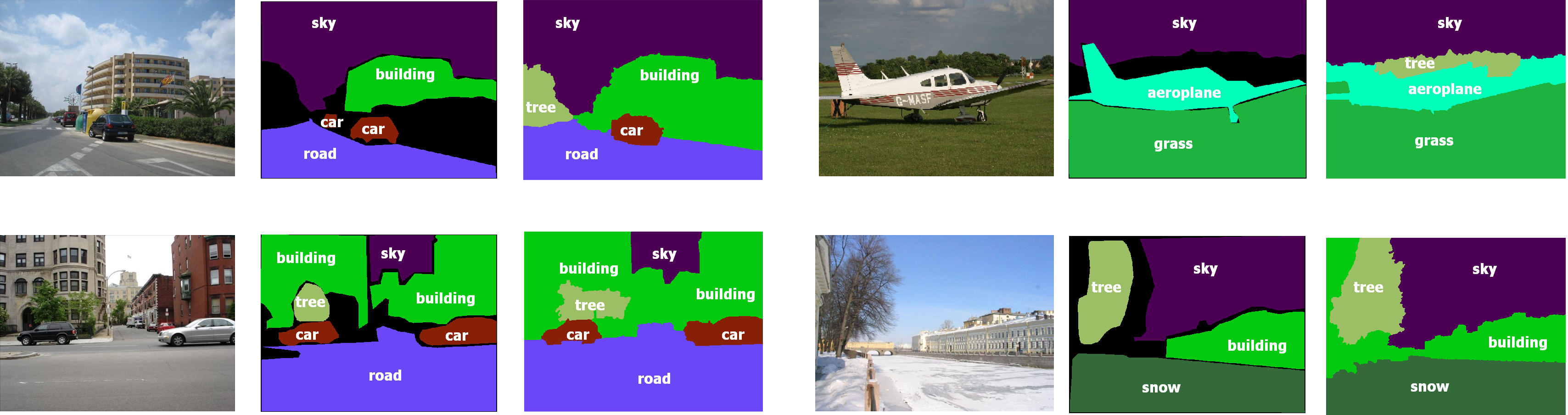}
\caption{Exemplar results on the 19-Category LabelMe dataset \cite{Jain}.  The test images, ground truth, and results from our proposed adaptive nonparametric method are shown in triple batches. Best viewed in color.}
\label{fig:others}

\end{figure*}

\textbf{How each new component affects SuperParsing \cite{TigheECCV2010} \ }  In order to study the impact of each newly proposed component, another experiment is conducted with different configuration settings. Namely, we report the results by incrementally adding LDA, super-pixel matching and adaptive nonparametric super-pixel classification to the traditional nonparametric image parsing pipeline~\cite{TigheECCV2010}, respectively. Keeping the fixed $k$ as 20 and the number of similar images in the retrieval set as 200, as recommended in \cite{TigheECCV2010} and adding LDA increase the performance of \cite{TigheECCV2010} by a small margin. We observe a large gain by adding super-pixel matching, \textit{i.e.}, 1.4\%, in per-pixel rate. Further adding adaptive nonparametric super-pixel classification drastically increases the combination (\cite{TigheECCV2010}, LDA, super-pixel matching, and fixed $k = 20$) by 1.1\% in per-pixel rate. For comparison, our work improves~\cite{TigheECCV2010} by \textbf{2.6\%} in terms of per-pixel rate and \textbf{5.2\%} in terms of per-category rate. The results clearly show the efficiency of our proposed super-pixel matching and adaptive nonparametric super-pixel classification. Figure \ref{fig:SIFTFlow_res} shows the exemplar results of different experimental settings on the SIFTFlow dataset. 

\begin{table}[!t]
\normalsize
\caption{The evaluation of the relevance of a retrieval set with respect to a query.}
\begin{center}
\begin{tabular}{|p{6.1cm} |c|}
\hline
\textbf{Retrieval Set Algorithm} & \textbf{~~NDCG~~}\\

\hline
GIST-based matching~\cite{LiuCe}& 0.83 \\
\hline
Global matching~\cite{TigheECCV2010}	& 0.85\\
\hline
Super-pixel matching	& 0.88\\
\hline
\end{tabular}
\end{center}

\label{table:ndcg}
\end{table}

\textbf{How good is the locality-aware retrieval set \ } We evaluate the performance of our retrieval set via Normalized Discounted Cumulative Gain (NDCG) \cite{NDCG} which is commonly used to evaluate ranking systems. NDCG is defined as follows, 
\begin{equation}
NDCG@k_r = \frac{1}{Z} \sum_{i=1}^{k_r}\frac{2^{rel(i)} - 1}{\log(i + 1)}, 
\end{equation}
where $rel(\cdot)$ is a binary value indicating whether the scene of the returned image is relevant (with value 1) or irrelevant (with value 0) to the one of the query image, and $Z$ is a constant to normalize the calculated score. Recall that $k_r$ is the number of returned images from locality-aware retrieval set to ensure the fair comparison. As shown in Table \ref{table:ndcg}, our super-pixel matching outperforms other baselines, namely, GIST-based matching and global matching in terms of NDCG. Figure \ref{fig:retrieval_res} also demonstrates the good results of our locality-aware retrieval set.

\textbf{Adaptive $k$ on different scene classes \ }Based on our hypothesis that the similar images should share the same $k$, we would like to study how the adaptive $k$ selection works for different types (scene classes) of similar images. To this end, we divide images in the SIFTFlow dataset into scene classes based on their filenames. For example, the test image ``coast\_arnat59.jpg'' is classified into \textit{coast} scene class. In total, there are 8 scene classes, namely, \textit{coast, forest, highway, inside city, mountain, open country, street}, and \textit{tall building}. We compute the mean number of categories (car, building, road, etc.) inside the testing set of the SIFTFlow dataset. Next, we compute the selected $k$ for each scene class by selecting the $k$ that has the highest confidence over all of the images in the same scene. The mean number of categories and the selected $k$ of each scene class are reported in Table \ref{table:Scene_K}.  As we can observe, the scene images with more object categories, i.e., highway, inside city and street, have lower $k$s. In contrast, the scene images with fewer object categories  have larger $k$s. Note that our method is unaware of the scene class of the test image. This means our method adapts well to different scene classes and brings the remarkable improvement to image parsing. In the preliminary experiments, we apply the randomization for the order of test super-pixels but the performance is similar to the one that is from $1$ to $n_q$. Therefore, the order of the super-pixels of test image does not affect the performance. 

\begin{table}[!t]
\normalsize

\caption{The mean number of categories and the correspondingly selected $k$ of each scene class on the SiftFlow dataset.}

\begin{center}
\begin{tabular}{|p{2.1cm} |c |c|}
\hline
\textbf{Scene Class} & \textbf{Mean No. of Categories} & \textbf{Selected $k$}\\

\hline
Coast& 3.8 &	12\\
\hline
Forest	& 2.5 &	36\\
\hline
Highway	& 6.5 &	6\\
\hline
Inside City	& 7.2 &	12\\
\hline
Mountain	& 2.6 &	22\\
\hline
Open Country	& 3.9 &	14\\
\hline
Street	& 7.5 &	6\\
\hline
Tall Building	& 3.3 &	43\\
\hline
\end{tabular}
\end{center}

\label{table:Scene_K}
\end{table}

\begin{table}[!t]
\normalsize
\caption{ Performance comparison of our algorithm with other algorithms on the 19-Category LabelMe dataset~\cite{Jain}. Per-pixel rates and average per-category rates are presented. The best performance values are marked in \textbf{bold}.}
\begin{center}
\begin{tabular}{"p{3.2cm} |c |c;}

\thickhline
\textbf{Algorithm} & \textbf{Per-Pixel (\%)} & \textbf{Per-Category (\%)}\\
\thickhline
\hline
\thinline
\multicolumn{3}{"c;}{Parametric Baselines} \\ 
\hline
Jain et al. \cite{Jain}	& 59.0	&	--	\\
\hline
Chen et al. \cite{ChenJain}	& 75.6	&	45.0\\

\thickhline
\hline
\thinline

\multicolumn{3}{"c;}{Nonparametric Baselines} \\ 

\hline
Myeong et al.~\cite{Heesoo2}	& 80.1	&	53.3\\
\hline

\multicolumn{3}{"c;}{Adaptive Nonparametric Algorithm} \\ 
\hline
Super-pixel Classification	& 80.3	&	53.3	\\
\hline
Contextual Smoothing 	& \textbf{82.7}	&	\textbf{55.1}\\
\thickhline
\end{tabular}
\end{center}

\label{table:Jain}
\end{table}

\subsection{Performance on 19-Category LabelMe Dataset}
Table \ref{table:Jain} shows the performance of our work compared with other baselines on the 19-Category LabelMe dataset. Our final adaptive nonparametric method on this dataset achieves \textbf{82.7\%}, surpassing all state-of-the-art performances. For the adaptive nonparametric method, our result has surpassed the one of Myeong et al. \cite{Heesoo2} by a large margin. Compared with the parametric method~\cite{ChenJain}, our work improves by \textbf{7.1\%}. Some exemplar results on this dataset are shown in Figure \ref{fig:others}.

\section{Conclusions and Future Work}
\label{sec:conclusion}
This paper has presented a novel approach to image parsing that can take advantage of adaptive nonparametric super-pixel classification. To the best of our knowledge, we are the first ones to exploit the locality-aware retrieval set and adaptive nonparametric super-pixel classification in image parsing. Extensive experimental results have clearly demonstrated the proposed method can achieve the state-of-the-art performance on diverse and challenging image parsing datasets. 



For future work, we are interested in exploring possible extensions to improve the performance. For example, the combination weight of different types of features can be learned. Another possible extension is to elegantly transfer other parameters apart from $k$, for example, the $\lambda$ of the contextual smoothing process from the retrieved training images to the test image. Since the current solution is specific for image parsing, we are also interested in generalizing the proposed method to other recognition tasks, such as image retrieval, and general $k$-NN classification applications. We also plan to leverage our work to video domain, i.e., action recognition~\cite{STAP} and human fixation prediction~\cite{TamMM}. 

Last but not least, to boost the super-pixel matching process, we can embed Locality-sensitive hashing (LSH) \cite{LSH} or the recently introduced Set Compression Tree (SCT) \cite{SCT} to encode the feature’s representative in few bits (instead of bytes) for large-scale matching. These coding methods and the insignificant number of super-pixels of each image make our super-pixel matching process feasible. In this paper, we only investigate the impact of adaptive non-parametric method in scene parsing. The utilization of LSH or SCT which are suitable for large-scale dataset will be considered for building a practical system in the future.

\bibliographystyle{IEEEtran}
\bibliography{tam2}

\begin{IEEEbiography}[{\includegraphics[width=1in,height=1.25in,clip,keepaspectratio]{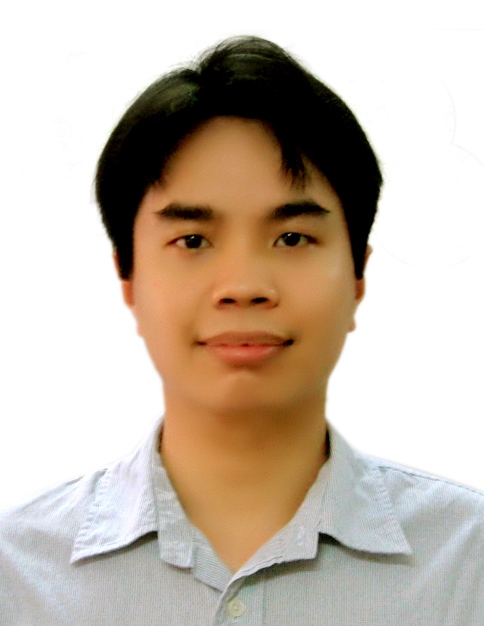}}]
{Dr. Tam V. Nguyen} obtained his Ph.D. degree with the Department of Electrical and Computer Engineering, National University of Singapore in 2013. Prior to that, he obtained his Master's degree from Chonnam National University, South Korea in 2009. He is currently a research scientist and principal investigator at the Applied Research and Technologies for Infocomm Centre at Singapore Polytechnic. His research interests include computer vision, machine learning and multimedia content analysis. He is the recipient of numerous awards including Young Vietnamese of the year 2005, the 2nd prize winner of ICPR 2012 contest on action recognition, and the best technical demonstration from ACM Multimedia 2012. 
\end{IEEEbiography}

\begin{IEEEbiography}[{\includegraphics[width=1in,height=1.25in,clip,keepaspectratio]{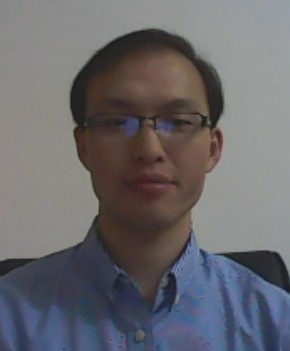}}]
{Canyi Lu} received the bachelor degree in mathematics from the Fuzhou University in 2009, and the master degree in the pattern recognition and intelligent system from the University of Science and Technology of China in 2012. He is currently a phd student with the Department of Electrical and Computer Engineering at the National University of Singapore. His current interests are the block diagonal affinity matrix learning and convex and nonconvex optimization. He was the winner of  2014 Microsoft Research Asia Fellowship. 
\end{IEEEbiography}
%

\begin{IEEEbiography}[{\includegraphics[width=1in,height=1.25in,clip,keepaspectratio]{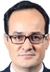}}]
{Dr. Jose Sepulveda} is the Director of the Applied Research and Technologies for Infocomm Centre at Singapore Polytechnic.  Dr Sepulveda's areas of interest include data analysis, technology in education.  Dr. Sepulveda's training background includes a PhD in Physics and an MBA. His postdoctoral research focused on bioinformatics at the Center for Genomics and Bioinformatics at Karolinska Institutet (Sweden) and at the Biochemistry Department at Baylor College of Medicine (USA).  After this, Dr Sepulveda went to work in technology in education and video game design (edutainment) at the Center for Technology in Teaching and Learning at Rice University in Houston.  His interests focuses on the use of technology to empower creativity and innovation (agile methods) to bridge the gap between ideas and commercial products, and the way that it can be used to support knowledge management and communities of practice.
\end{IEEEbiography}


\begin{IEEEbiography}[{\includegraphics[width=1in,height=1.25in,clip,keepaspectratio]{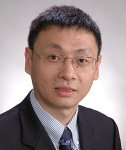}}]
{Dr. Shuicheng Yan}
(M'06, SM'09)  is currently an Associate Professor at the Department of Electrical and Computer Engineering at National University of Singapore, and the founding lead of the Learning and Vision Research Group (http://www.lv-nus.org). Dr. Yan's research areas include machine learning, computer vision and multimedia, and he has authored/co-authored hundreds of technical papers over a wide range of research topics, with Google Scholar citation $>$14,000 times and H-index 52. He is ISI Highly-cited Researcher, 2014 and IAPR Fellow 2014. He has been serving as an associate editor of IEEE TKDE, TCSVT and ACM Transactions on Intelligent Systems and Technology (ACM TIST). He received the Best Paper Awards from ACM MM'13 (Best Paper and Best Student Paper), ACM MM’12 (Best Demo), PCM'11, ACM MM'10, ICME'10 and ICIMCS'09, the runner-up prize of ILSVRC'13, the winner prize of ILSVRC'14 detection task, the winner prizes of the classification task in PASCAL VOC 2010-2012, the winner prize of the segmentation task in PASCAL VOC 2012, the honourable mention prize of the detection task in PASCAL VOC'10, 2010 TCSVT Best Associate Editor (BAE) Award, 2010 Young Faculty Research Award, 2011 Singapore Young Scientist Award, and 2012 NUS Young Researcher Award.
\end{IEEEbiography}

\vfill



\end{document}